\documentclass{article}
\usepackage{spconf,amsmath,graphicx}
\usepackage[utf8]{inputenc} 
\usepackage[T1]{fontenc}    
\usepackage{hyperref}       
\usepackage{url}            
\usepackage{booktabs}       
\usepackage{amsfonts}       
\usepackage{nicefrac}       
\usepackage{microtype}      
\usepackage{graphicx}
\usepackage{amsmath,amssymb,amsthm}
\usepackage{breqn}
\usepackage[ruled,vlined]{algorithm2e}
\usepackage[utf8]{inputenc}
\usepackage[english]{babel}
\newtheorem{theorem}{Theorem}
\graphicspath{ {./images/} } 
\usepackage{setspace}
\usepackage{mathtools}
\usepackage[title]{appendix}
\usepackage{color}
\usepackage{multirow}
\usepackage{tabularx}

\newtheorem{claim}{Claim}

\title{Refining Self-Supervised Learning in Imaging: Beyond Linear Metric}
\name{Bo Jiang$^{\star}$\thanks{Thanks to the generous support of ARO grant W911NF1910202.} \qquad Hamid Krim$^{\star}$ \qquad Tianfu Wu$^{\star}$ \qquad Derya Cansever$^{\dagger}$}
  
\address{$^{\star}$ North Carolina State University \\
    $^{\dagger}$US Army Research Office}
\begin{document}
%
\maketitle
\begin{abstract}
We introduce in this paper a new statistical perspective, exploiting the Jaccard similarity metric, as a measure-based metric to effectively invoke non-linear features in the loss of self-supervised contrastive learning. Specifically, our proposed metric may be interpreted as a dependence measure between two adapted projections learned from the so-called latent representations. This is in contrast to the cosine similarity measure in the conventional contrastive learning model, which accounts for correlation information. To the best of our knowledge, this effectively non-linearly fused information embedded in the Jaccard similarity, is novel to self-supervision learning with promising results. The proposed approach is compared to two state-of-the-art self-supervised contrastive learning methods on three image datasets. We not only demonstrate its amenable applicability in current ML problems, but also its improved performance and training efficiency. 
\end{abstract}
\section{Introduction}
The notion of latent representation of data has, over the last few years, emerged as a significant catalyst\cite{jiang2021dynamicjournal}, particularly in computer vision, where a visual representation\cite{jiang2021dynamic,tang2016analysis,tang2018structured} in the latent space provides a powerful grounding for supervised learning and inference. Scaling learning-based inference to large datasets (e.g., ImageNet), however, is labor-intensive and time-consuming, and calls for a more viable alternative\cite{tang2021deep,tang2022deep}. The increasing popularity of Self-Supervised Learning (SSL) particularly in Computer Vision (CV) in conjunction with learned visual representations \cite{chen2020simple,he2020momentum} and pretext tasks \cite{doersch2015unsupervised,jenni2018self,kim2018learning} seeks labelling from a diverse view or a fraction of a given data sample. A key feature of SSL is the adopted contrastive loss, which is aimed at discriminating positive keys (i.e., different views) from negative ones in the latent space of images. 

To define some notion of semantic space, a contrastive learning model was proposed in \cite{chopra2005learning}. In so doing, they proposed a $L_2$-norm distance to estimate similarity among two images\cite{hamza2003jensen}. This procedure entailed minimizing a semantic distance between images from the same class while maximizing that from different classes. 
The recent focus in research has gradually shifted to only minimizing the distance between the positive pairs without simultaneously maximizing the distance between negative pairs \cite{chen2021exploring,grill2020bootstrap,tian2021understanding}.  Given different views of the original images, the encoder is trained by minimizing the distance between different views in the hidden space. The paper\cite{tian2021understanding} also highlights the impact of a potential asymmetry with the larger number of negative keys than the number of positive keys skewing the learning.
Empirically choosing the transformation and augmentation methods\cite{chen2020big} and handling negative samples\cite{chuang2020debiased} may also induce biases and hence significantly affect the learning encoders' performance.
As a result of these augmentation procedures, a selection of a correct metric is in order as it will clearly impact any training procedure. The adoption of a cosine similarity implies a linear measurability of the similarity in the latent space. The features are, however, the result of non-linear processing across the learning layers and are hence elements lying on some manifold rather than on a linear space. To that end, we propose to account for the intrinsic non-linearity by proposing the Jaccard similarity coefficient, developed in the early 1900's\cite{jaccard1912distribution}, to capture non-linear characteristics in the similarity between finite sets of prevailing features. Upon interpreting SSL as an inference problem, and using a hypothesis test for a quantitative decision in contrastive learning, we propose a bi-projector Jaccard-based loss which effectively reflects the non-linearities in the hidden spaces. 

Our proposed adaptive approach effectively aims at addressing the inherent modal structure of the latent/pseudo-invariant characteristics of the data with a comparison to two well-known self-supervised learning models SimCLR\cite{chen2020simple} and MoCo\cite{he2020momentum}, based on both the k-nearest neighbors (k-NN) and a standard linear classification methods. The selected datasets are CIFAR-10, CIFAR-100, and Tiny-ImageNet-200 to evaluate the performance of SSL.
Our contributions in this paper include (1) Our providing a theoretically sound and tractable hypothesis testing formulation of the contrastive loss, and of the ensuing projection for procedure stabilization; (2) Our introduction of a bi-projector model along with a novel Jaccard-similarity-based contrastive loss evaluation; (3) Our derivation of similarity maximization/minimization of positive/negative pairs in each of the hidden spaces separately accounting for the non-linear features by a Jaccard metric-based fusion.

\section{SSL: Background and Development}
\subsection{Notation}
Throughout the paper, we will adopt a notationally caligraphic upper case letter to denote a space, and an upper case letter to denote a data representation in different spaces, specifically, $X$ represents the original data, $X^K$ the transformation and augmentation of the original image data in the pixel space, with $Q$ and $K$ respectively denoting the features in the latent representation spaces of the original and augmented images. We use the subscripted lower case letter to distinguish different elements in different spaces, i.e., $x_i$ indicates the $i^{th}$ object element of $X$. We consider $f_\theta(\cdot)$ to be a function parameterized by $\theta$.
\subsection{Contrastive Loss}
Given an unlabeled dataset $X=\{x_1,x_2,\dots,x_N\}$, a similarity calculation method $s(\cdot,\cdot)$, an embedding function $f_{\theta}(\cdot)$, and a projector $g_{\theta_p}(\cdot)$, which collaboratively projects the data from the original space $\mathcal{X}$ into a flat hidden space $\mathcal{H}$. Define $x$ a data sample in  $X$, $x^+$ as a positive key, and $x_j^-$ denotes the $j^{th}$ of negative keys of $x$, $h_{q}=g_{\theta_p}\circ f_\theta(x)$, $h_{k^+}=g_{\theta_p}\circ f_\theta(x^+)$, and $h_{k_j^-}=g_{\theta_p}\circ f_\theta(x_j^-)$. To carry through the original idea of contrasting positive with negative keys, a triplet loss, was first proposed in \cite{schroff2015facenet,hadsell2006dimensionality}, and is aimed at learning an invariant mapping to project high dimensional features onto a low dimensional space, such that intra-class features are close to each other while keeping inter-class features far apart,
$\mathcal{L}_{triplet}(x, x^+,x^-|\theta,\theta_p)
=\mathbf{max}(0,d(h_q,h_{k^+})-d(h_q,h_{k^-})+m)$,
with $d(\cdot,\cdot)$ denoting a metric, and $m$ is some margin that restricts the largest meaningful distance difference between intra-class and inter-class latent representations. A probabilistic interpretation has recently emerged "softmax" expression to highlight the frequential aspect of sampling,
$\mathcal{L}_{co}(x,x^{+},x_j^{-})
=-\mathbf{log}\bigg( \frac{\exp\big(s(h_{q},h_{k^+})/\tau\big)} {\exp\big(s(h_{q},h_{k^+})/\tau\big)+\sum_j \exp\big(s(h_{q},h_{k_j^-})/\tau\big)} \bigg)$,
where $\tau$ is referred to as the temperature scalar in the contrastive loss to modulate the influence of the exponent term. Note that the cosine similarity expression $\operatorname{Cos-Sim}(x,y)~=~\frac{\langle x,y\rangle}{\|x\|\|y\|}$ is a common similarity measure choice between latent representations. While seemingly distinct, these two losses bear a relation as shown next,

\begin{theorem}
 The mean zero-margin triplet loss is upper bounded and  smoothly approximated by the contrastive loss.
\end{theorem} 

\noindent\textbf{Proof:} The LogSumExp term in the denominator can be interpreted as the smooth approximation to the maximum function. Consider a finite set $Y~=~\{y_1,...,y_n\}$, $\mathbf{log}\bigg(e^{\mathbf{max}(Y)}\big)\bigg)\leq\mathbf{log}\bigg(\sum_i e^{y_i}\bigg)\leq\mathbf{log}\bigg(n e^{\mathbf{max}(Y)}\bigg)$.
While the so-called soft-max is often applied in ML to avoid non-smoothness when calculating the gradient, it often is implicit in other settings.  The contrastive loss equation thus yields the mean triplet loss with the zero margin. By defining $s^+=\operatorname{Cos-Sim}(h_q,h_{k^+})/\tau$ and $s_j^-=\operatorname{Cos-Sim}(h_q,h_{k_j^-})/\tau$, we can rewrite,
\begin{equation}
\begin{split}
    \mathcal{L}_{co}& = -\mathbf{log}\bigg(\frac{e^{s^+}}{e^{s^+}+\sum_i e^{s_j^-}}\bigg) = \mathbf{log}\bigg(e^0+\sum_i e^{s_j^--s^+}\bigg)\\
    & \geq \mathbf{max}\bigg(0,s^-_1-s^+,\dots,s^-_n-s^+\bigg) \geq~\frac{1}{n}\sum_j\mathcal{L}_{triplet}^j.
\end{split}
\end{equation}
The minimization of the contrastive loss function implies a smooth enhancement of proximity of positive pairs and of separation of negative pairs.\hfill\qedsymbol

\subsection{A Variation on a Theme: A Jaccard-index Assessment}
\subsubsection{Jaccard Similarity}
The Jaccard index\cite{jaccard1912distribution}, commonly called the Jaccard similarity coefficient, is a set-theoretic measure to generically evaluate similarity between two finite sample sets.  In general, the Jaccard similarity coefficient is evaluated as a ratio of the intersection of two finite sets $A$ and $B$ and their union,
$J(A,B)~=~\frac{|A\cap B|}{|A\cup B|}~=~\frac{|A\cap B|}{|A\setminus B|+|B\setminus A|+|A\cap B|}$.
An alternative measure-inspired interpretation of the Jaccard index may also be attained. Specifically, an information theoretic/probabilistic interpretation unveils a measure-theoretic hence deeper comparison using a probability measure $\mu$ on the  measurable sample space $S$, yielding a similar coefficient as, 
$J_\mu(A,B)~=~\frac{\mu(A\cap B)}{\mu(A\cup B)}$.

The use of measures (i.e., a probabilistic model capturing all moments), particularly points to the significance of the information scoped out and beyond the linear (i.e., second-order moment) typically invoked by a Cosine similarity. This is hence consistent with our goal of gleaning additional secondary information beyond that normally sought in the latent space of the classes. 
The flexibility of the proposed Jaccard similarity measure accounts for the inherent nonlinear information moments, thus affording a more accurate assessment of non-linear features in the latent space.

\subsubsection{Proposed Contrastive Learning  Model}
A hypothesis is a proposed explanation of an observed phenomenon\cite{lehmann2006testing}.
Self-supervised contrastive learning, viewed as a guessing exercise, is the assessment result of randomly sampled data and its subsequent assignment to either a positive or negative class, each with an underlying distribution. 
As a result, a null hypothesis $H_0$ stating that two observations (i.e., images) are from the same class, with a corresponding distributional score  $f_{\theta^+}(\cdot,\cdot)$. Similarly, an  alternative hypothesis $H_1$ is defined for two observations  from different classes, with a corresponding score $f_{\theta^-}(\cdot,\cdot)$.

With the inferential perspective in mind and the Jaccard similarity measure in hand, we propose to seek  hidden representations as finite information sets/distributions of original objects. As is typically interpreted with measures in probabilistic settings, intersections will account for similarity, while set differences reflect dissimilarities (disparities) among the data samples. To that end, we introduce a bi-projector system model $g_{\theta_i}(\cdot), i=1, 2$, with the first following $H_0$ (i.e., that two objects are from the same class by measuring the intersection of extracted features), while the other following $H_1$. We can hence define two hypotheses scoring functions, 
$f(x_1,x_2|H_0,\theta)~\coloneqq~f_{\theta^+}(\cdot,\cdot)~=~s\circ g_{\theta_1}\circ f_\theta,$ and $f(x_1,x_2|H_1,\theta)~\coloneqq~f_{\theta^-}(\cdot,\cdot)~=~ds\circ g_{\theta_2}\circ f_\theta,$
respectively, where $s(\cdot,\cdot)$,$ds(\cdot,\cdot)$ is each respectively, a similarity and dissimilarity measure.
\begin{figure}
  \centering
  \includegraphics[scale = 0.18]{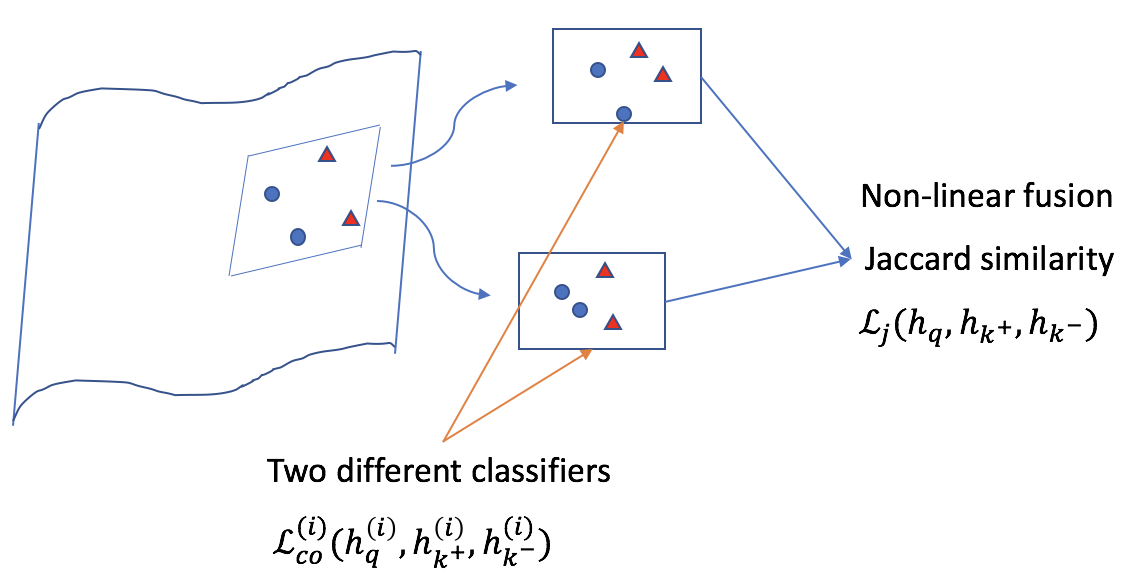}
  \caption{An intuitive explanation of our model.}
  \label{fig:model_intuition}
\end{figure}
Fig.~\ref{fig:model_intuition} provides a high-level and  intuitive description of the proposed rationale. 
The latent representations of the original data lying on an unknown curved manifold, make it unreasonable and possibly detrimental to the training/learning, to be using Euclidean-based similarity metrics. The key idea is to then learn a homeomorphism $g_\theta(\cdot)$ that projects the latent space information onto a set of Euclidean spaces, where a similarity can be measured by $L^2$-norm, and consistently integrate the extracted information. In our case, we hence seek two distinct projectors which directly contribute to the computation of the Jaccard similarity which is in turn, used to optimize the encoder. A carefully designed combination of the information extracted by the two projectors as described next, will yield a Jaccard similarity.

\subsubsection{Proposed Algorithmic Solution}
This Jaccard-based similarity/loss provides a non-linear combination (fusion) of features thereby avoiding asymmetry in the training. Different measurement focus also prevent the two projectors from collapsing into one single point. To proceed with the detailed development of the Jaccard index-based model, we first note that the similarity/dissimilarity are Euclidean-based (i.e., $ds(x,y)=\parallel x-y\parallel^2$), where $x, y$ are normalized in the hidden space. Specifically, for given latent representations of two images $q_1$ and $q_2$, we only consider the two finite information sets extracted by the first projector $g_{\theta_1}(\cdot)$ for measuring the intersection, $\langle g_{\theta_1}(q_1),g_{\theta_1}(q_2)\rangle$, while the second projector is for measuring the dissimilarity, $\|g_{\theta_2}(q_1)-g_{\theta_2}(q_2)\|_2^2$, when ultimately computing the Jaccard similarity.

\begin{claim} 
The Jaccard index, acting on two hidden spaces to measure their similarity is equivalent to respectively maximizing and minimizing the Jaccard similarity between positive and negative pairs.
\begin{equation}
\begin{split}
    \mathcal{L}_{J-Tri}({\mathbf a})~&=~-\left.\frac{\mu(x\cap x^+)}{\mu(x\cup x^+)}+\frac{\mu(x\cap x_i^-)}{\mu(x\cup x_i^-)}\right\vert_{\theta,\theta_1,\theta_2}\\
    & = \frac{s^+}{s^++ds^+}-\frac{s_i^-}{s_i^-+ds_i^-}.
\end{split}
\end{equation}
where ${\mathbf a}=(x,x^+,x_i^-,\theta,\theta_1,\theta_2) $, $s^+~=~\langle g_{\theta_1}\circ f_\theta(x),g_{\theta_1}\circ f_\theta(x^+)\rangle$, $s_i^-~=~\langle g_{\theta_1}\circ f_\theta(x),g_{\theta_1}\circ f_\theta(x_i^-)\rangle$, $ds^+~=~\|g_{\theta_2}\circ f_\theta(x)-g_{\theta_2}\circ f_\theta(x^+)\|^2_2$, and $ds_i^-~=~\|g_{\theta_2}\circ f_\theta(x)-g_{\theta_2}\circ f_\theta(x_i^-)\|^2_2$.
\end{claim}
In Eq.(1), we prove that the contrastive loss is a smooth approximation to the triplet loss. Therefore, we can also apply the smoothness property of LogSumExp function to Jaccard-based triplet loss. Then, the smooth version of Jaccard-based contrastive loss is shown below,
\begin{equation}
\begin{split}
    \mathcal{L}_{J-Tri}(a) & = -\frac{1}{N}\sum_i\bigg(\frac{s^+}{s^++ds^+}-\frac{s_i^-}{s_i^-+ds_i^-}\bigg)\\
    & \approx-\mathbf{log}\bigg(\frac{e^{\frac{s^+}{s^++ds^+}}}{e^{\frac{s^+}{s^++ds^+}}+\sum_ie^{\frac{s_i^-}{s_i^-+ds_i^-}}}\bigg).
\end{split}
\end{equation}

To better achieve the intra/inter-class clustering in the hidden spaces, we follow the conventional self-supervised learning and calculate the contrastive loss among positive and negative keys in each hidden space, respectively. Subsequently applying the Jaccard similarity to achieve the feature fusion between two hidden spaces (as noted with sensitivity to similarity and dissimilarity, respectively) yields the proposed model detailed next. As a result we postulate the following,

\noindent\textbf{Proposed self-supervision loss:}
\begin{equation}
\begin{split}
    & \mathcal{L}_J(a) =-\alpha_1\mathbf{log}\bigg(\frac{e^{\langle h_{q}^{(1)},h_{k^+}^{(1)}\rangle/\tau}}{e^{\langle h_{q}^{(1)},h_{k^+}^{(1)}\rangle/\tau}+\sum_i e^{\langle h_{q}^{(1)},h_{k_i^-}^{(1)}\rangle/\tau}}\bigg)\\
    & -\alpha_2\mathbf{log}\bigg(\frac{e^{\langle h_{q}^{(2)},h_{k^+}^{(2)}\rangle/\tau}}{e^{\langle h_{q}^{(2)},h_{k^+}^{(2)}\rangle/\tau}+\sum_i e^{\langle h_{q}^{(2)},h_{k_i^-}^{(1)}\rangle/\tau}}\bigg)\\
    & -(1-\alpha_1-\alpha_2)\times\\
    & {\scriptstyle \mathbf{log}\bigg(\frac{e^{\frac{\langle h_{q}^{(1)},h_{k^+}^{(1)}\rangle}{\langle h_{q}^{(1)},h_{k^+}^{(1)}\rangle+\|h_{q}^{(2)}-h_{k^+}^{(2)}\|^2_2}/\tau}}{e^{\frac{\langle h_{q}^{(1)},h_{k^+}^{(1)}\rangle}{\langle h_{q}^{(1)},h_{k^+}^{(1)}\rangle+\|h_{q}^{(2)}-h_{k^+}^{(2)}\|^2_2}/\tau}+\sum_i e^{\frac{\langle h_{q}^{(1)},h_{k_i^-}^{(1)}\rangle}{\langle h_{q}^{(1)},h_{k_i^-}^{(1)}\rangle+\|h_{q}^{(2)}-h_{k_i^-}^{(2)}\|^2_2}/\tau}}\bigg)},
\end{split}
\end{equation}
where $h_q^{(j)}~=~g_{\theta_j}\circ f_{\theta}(x)$, $h_{k^+}^{(j)}~=~g_{\theta_j}\circ f_{\theta}(x^+)$, and $h_{k_i^-}^{(j)}~=~g_{\theta_j}\circ f_{\theta}(x_i^-)$, $j=1,2$.

\section{Experimental Results}
In this section, we demonstrate the compatibility, efficiency, and accuracy of the Jaccard-similarity-based model. We consider SimCLR\cite{chen2020simple} and MoCo\cite{he2020momentum} as the benchmarks and test the models' performance on CIFAR-10/100\cite{krizhevsky2009learning} and Tiny-imagenet-200\cite{le2015tiny}. Due to our limited GPU resources, we can not run full-scale experiments on the ImageNet-1k.

In the SSL training stage, we use ResNet-50\cite{he2016deep} as the backbone network for all models. All neural networks are trained with a batch size of 512 for 500 epochs on both CIFAR-10 and CIFAR-100 datasets and a batch size of 256 on the Tiny-Imagenet dataset. All models apply the same set of augmentation methods as SimCLR, and a constant learning rate $10^{-3}$.
We fix the encoder after each training epoch and apply the k-NN classification method to verify the encoder's performance during the training. Different temperature parameters ($\tau$) are chosen for training, and we choose the best encoder's performance as the model's final result. We have reproduced the results of the two benchmarks.

\begin{figure}[h]
  \centering
  \includegraphics[width=6.8cm, height = 6.5cm]{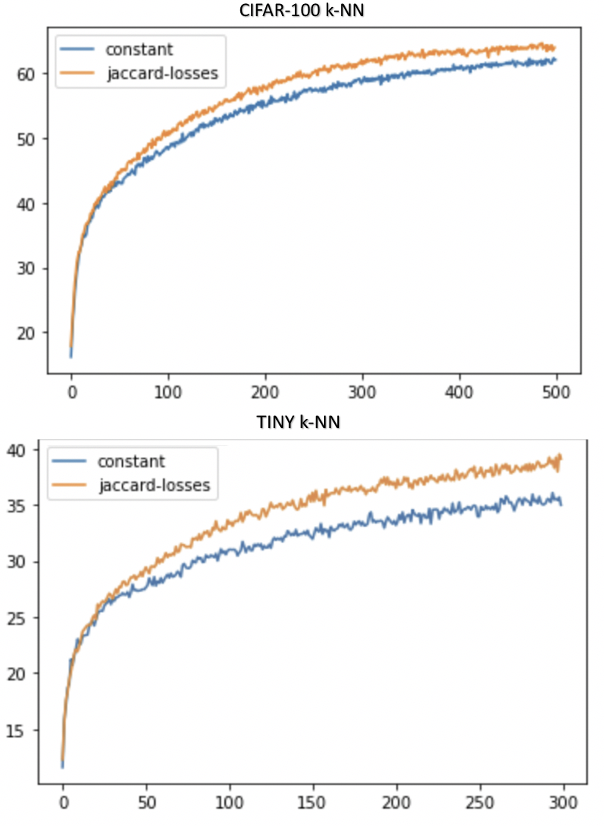}
  \caption{k-NN accuracy}
  \label{fig:k_NN_acc}
\end{figure}

We show in Table \ref{tab:computational_complexity} the complexity, i.e., the running time, of our model compared with SimCLR. The training time of one epoch is recorded on one NVIDIA Tesla V100 GPU for all models. Only one additional projector, a 2-layer fully connected neural network, is added to our model, and the loss is computed in matrix format. The recording time shows our model takes 1.1 times of SimCLR running time in CIFAR datasets and 1.01 times of it in the Tiny-Imagenet-200 dataset. Fig.\ref{fig:k_NN_acc} shows the encoders' performance curves of SimCLR and our model based on the k-NN method w.r.t the training epoch. We can observe that our model's encoder demonstrates quite a convincing learning speed compared with the conventional model with the same learning rate setup.
\begin{table}[h]
\renewcommand{\arraystretch}{1}
    \centering
    \begin{tabular}{|c|c|c|}
    \hline
    Dataset & CIFAR & Tiny-Imagenet-200\\
    \hline
    SimCLR &  1m36s/epoch & 13m51s/epoch\\
    \hline
    Our model & 1m47s/epoch & 14m01s/epoch\\
    \hline
    \end{tabular}
    \caption{Running time.}
    \label{tab:computational_complexity}
\end{table}

Table \ref{tab:our_model} shows the linear evaluation results' comparison on different datasets between conventional self-supervised learning models and our modified models. Note that compared with the conventional contrastive learning model, the complexity of the whole system is rarely increased, and the main difference is based on the similarity metric. The results show that our proposal of considering a non-linear metric helps improve the accuracy in most comparable experiments. The more complex the data is, the more improvement is achieved, confirming that the non-linear impart is more pronounced. These observations reflect the fact that latent representations of a complex dataset lie on non-linear manifold, where the linear metric may lead to imprecise estimation. Overall, the encoder based on SimCLR setup with our proposed model obtains the best performance in all datasets.

\begin{table}[h]
\renewcommand{\arraystretch}{1}
    \centering
    \begin{tabular}{|c c||c|c|c|}
    \hline
    \multicolumn{2}{|c|}{Top-1 Accuracy (\%)} & \multicolumn{3}{|c|}{Dataset}\\
    \hline
    \multirow{2}{*}{Model} &\multirow{2}{*}{Method}  & CIFAR & CIFAR & Tiny\\
     &  & 10    & 100   & ImageNet\\
    \hline
    \hline
    \multirow{2}{*}{SimCLR} & k-NN & 88.6 & 62.1 & 35.9\\
    & linear & 91.5 & 68.6 & 50.7\\
    \hline
    SimCLR + & k-NN & 88.6 & 63.9 & 39.5\\
    our model & linear & 91.8 & 69.5 & 53.8\\
    \hline
    \hline
    \multirow{2}{*}{MoCo} & k-NN & 85.5 & 56.8 & 33.5\\
    & linear & 90.5 & 66.8 & 49.6\\
    \hline
    MoCo + & k-NN & 85.1 & 59.2 & 35.4\\
    our model & linear & 90.2 & 67.5 & 51.2\\
    \hline
    \end{tabular}
    \caption{Linear evaluation results.}
    \label{tab:our_model}
\end{table}

\section{Conclusions}
This paper explains the contrastive loss from a hypothesis-testing point of view by distinguishing whether a pair of images is from the same category. We subsequently introduce a bi-projector model and a Jaccard-similarity-based loss to fuse the information extracted by the two projectors. In the experiments, we demonstrate the compatibility of our model by applying it as a plug-in unit to the state-of-art approaches. We also show that our model outperforms state-of-art contrastive SSL methods. The accuracy improvement is more pronounced for more complex datasets proving that the similarity measure should account for non-linearity on the latent space. Future research can consider model transport between the samples, to provide better insight on their behavior on the manifold.

\newpage
\bibliographystyle{unsrt}
\bibliography{reference.bib}

\end{document}